\documentclass[conference]{IEEEtran}
\IEEEoverridecommandlockouts
\usepackage{cite}
\usepackage{amsmath, amssymb, amsfonts}
\usepackage{algorithmic}
\usepackage{graphicx} 
\usepackage{textcomp}
\usepackage{xcolor}
\usepackage{comment}
\usepackage{colortbl}
\usepackage{dhucs} 
\usepackage{booktabs} 
\usepackage{multirow}
\usepackage{setspace} 
\usepackage{makecell} 
\usepackage{enumitem} 
\usepackage{nicematrix} 
\usepackage{threeparttable} 
\usepackage{dblfloatfix} 
\usepackage[ruled,vlined,algo2e]{algorithm2e}

\usepackage{lineno}
\usepackage{relsize}
\usepackage{tabularx}
\usepackage{diagbox}
\usepackage{hhline}
\usepackage{bm}
\usepackage{array}
\usepackage{mathtools}
\usepackage{tabularx,array,hhline}
\usepackage{subcaption} 
\usepackage{array,booktabs,tabularx}
\usepackage{tikz}
\usetikzlibrary{calc}

\usepackage[font=small,labelfont=small]{caption}

\newcommand\refFigure[1]{Fig.~\ref{#1}}
\newcommand\refTable[1]{Table~\ref{#1}}
\newcommand\refSection[1]{Section~\ref{#1}}
\newcommand\refAlgo[1]{Algorithm~\ref{#1}}
\newcommand\refEqn[1]{(\ref{#1})}

\newcommand\korean[1]{}
\newcommand\blind[1]{XXXX}

\def\useshortref{0}
\def\usemiddleref{0}
\if\useshortref 1

\else

\if\usemiddleref 1

\else

\fi
\fi

\if\useshortref 1

\else

\if\usemiddleref 1

\else

\fi
\fi

\usepackage[a4paper, total={184mm,243mm}]{geometry}
\def\BibTeX{{\rm B\kern-.05em{\sc i\kern-.025em b}\kern-.08em
    T\kern-.1667em\lower.7ex\hbox{E}\kern-.125emX}}

\author{
Eun-Su Cho$^{ \dag \sharp}$, Jongin Choi$^{  \dag \sharp}$, Jeongmin Jin$^{  \dag}$, Jae-Jin Lee$^{\ddag}$ and Woojoo Lee$^{ \dag  \ast}$\\
$^{ \dag}$School of Intelligent Semiconductor Engineering, Chung-Ang University, Seoul, Korea \\
$^{ \ddag}$AI Edge SoC Research Section, the Electronics and Telecommunications Research Institute, Deajeon, Korea \\

\thanks{
This paper has been accepted for publication at the Design, Automation and Test in Europe (\textit{DATE 2026}). 
This document represents the camera-ready version.

$^{\sharp}$ Eun-Su Cho and Jongin Choi contributed equally to this work.

$^{*}$Woojoo Lee is the corresponding author.}
}

\begin{document}

\title{\LARGE\relsize{+1.5} FiCABU: A Fisher‑Based, Context‑Adaptive Machine Unlearning Processor for Edge AI\\
}

\IEEEaftertitletext{\vspace{-2ex}}

\maketitle


\begin{abstract}


Machine unlearning, driven by privacy regulations and the “right to be forgotten,” is increasingly needed at the edge, yet server-centric or retraining-heavy methods are impractical under tight computation and energy budgets. 
We present \emph{FiCABU} (Fisher-based Context-Adaptive Balanced Unlearning), a SW–HW co-design that brings unlearning to edge AI processors. 
FiCABU combines (i) \emph{Context-Adaptive Unlearning}, which begins edits from back-end layers and halts once the target forgetting is reached, with (ii) \emph{Balanced Dampening}, which scales dampening strength by depth to preserve retain accuracy. 
These methods are realized in a full RTL design of a RISC-V edge AI processor that integrates two lightweight IPs for Fisher estimation and dampening into a GEMM-centric streaming pipeline, validated on an FPGA prototype and synthesized in 45\,nm for power analysis. 
Across CIFAR-20 and PinsFaceRecognition with ResNet-18 and \emph{ViT}, FiCABU achieves random-guess forget accuracy while matching the retraining-free Selective Synaptic Dampening (SSD) baseline on retain accuracy, reducing computation by up to 87.52\% (ResNet-18) and 71.03\% (\emph{ViT}). 
On the INT8 hardware prototype, FiCABU further improves retain preservation and reduces energy to {6.48\%} (CIFAR-20) and {0.13\%} (PinsFaceRecognition) of the SSD baseline. 
In sum, FiCABU demonstrates that back-end–first, depth-aware unlearning can be made both practical and efficient for resource-constrained edge AI devices.

\end{abstract}

\section{Introduction}
\vspace{-2pt}
As machine learning models are deployed across increasingly diverse domains, the scale of training corpora continues to grow~\cite{Yun:DATE2025}. These corpora often contain sensitive personal information (e.g., biometrics), and legal as well as societal pressures around the “right to be forgotten” are mounting~\cite{Voigt:GDPR2017, Goldman:2020}. 
Beyond removing samples from a dataset, a deployed model must behave \emph{as if} specified data never influenced its parameters. 
This motivates \emph{machine unlearning}—removing the effect of designated data from a pre-trained model at the parameter level~\cite{Huang:IOT2025}. 
Unlearning is valuable not only for compliance but also for maintaining model quality by attenuating the influence of erroneous or stale information~\cite{Xu:ETCI2024, Kurmanji:NIPS2023}.

Most prior work studies unlearning under server-class assumptions—data collected on edge devices (e.g., IoT sensors, embedded cameras, or wearable health monitors) are uploaded, and unlearning is executed in the cloud~\cite{Hu:CCS2024}. However, edge compute capabilities have improved substantially~\cite{Xia:SP2025}, and—critically—the party that collects the data to be deleted is often the edge device itself. This raises a natural question: should unlearning be performed \emph{on-device} instead of in the cloud? 
As illustrated in \refFigure{fig:intro}, cloud-based unlearning incurs communication latency/energy overheads and additional privacy exposure, whereas executing unlearning where the data originate mitigates both concerns at their source~\cite{Bayerl:DATE2020}. 
At the same time, edge platforms remain resource-constrained relative to servers; thus, a practical on-device solution must be carefully tailored to edge budgets. 
Inference on the edge has long embraced \emph{adaptive/selective processing}—doing useful work while avoiding unnecessary computation~\cite{Motetti:DATE2024, Korol:DATE2023}. We argue that on-device unlearning should adopt a similar ethos.

\begingroup
\begin{figure}[t]
\centering
\includegraphics[width=0.86\columnwidth]{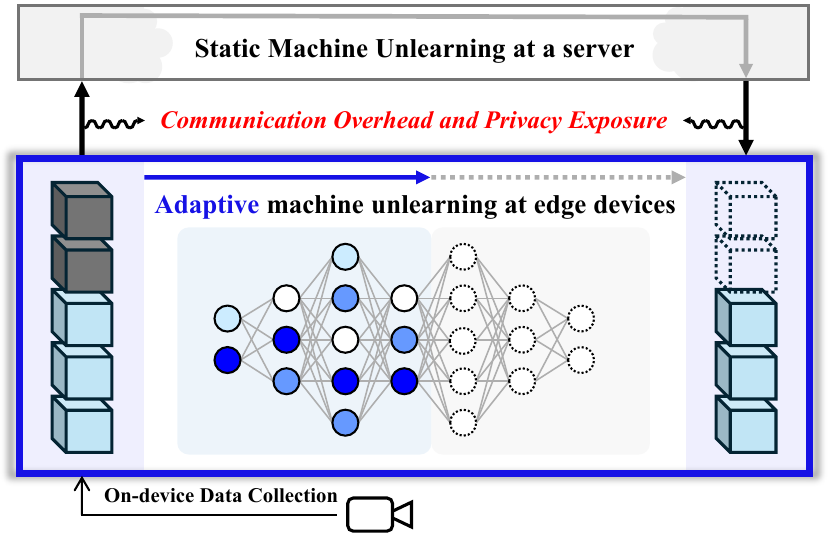}
\vskip -4pt
\caption{\small Comparison of server-based unlearning, which incurs communication and privacy costs, and adaptive on-device unlearning, which performs selective updates locally.}
\label{fig:intro}
\vskip -4pt
\end{figure}
\endgroup

We therefore begin by examining algorithmic options for unlearning under edge constraints. 
The most definitive approach—retraining a model from scratch after excising the forget samples—guarantees removal but is prohibitively expensive for modern models~\cite{yao:ACL2024}. 
Partitioned training lowers some costs yet still depends on (re)training~\cite{Bourtoule:SP2021}. 
Structure- or precision-oriented edits (e.g., pruning, quantization) simplify parameters encoding fine-grained characteristics of the forget set, but typically require additional training to restore utility and remain heavyweight on the edge~\cite{Jia:NIPS2023, tong:ICCV2025, Tu:CV2025, wang:CV2025}. 
More recently, retraining-free methods estimate parameter importance and update only those most attributable to the forget set. 
A prominent line leverages the diagonal approximation of the Fisher Information Matrix (\textit{FIM}), which estimates sensitivity from squared gradients in a single forward–loss evaluation, avoiding repeated dataset passes~\cite{Foster:AAAI2023}. 
This direction progressed from early ad hoc formulations~\cite{guo:arxiv2023} to post-hoc edits on pre-trained models~\cite{Golatkar:CVPR2020} and zero-shot variants~\cite{Sekhari:NIPS2021}, alongside recent gradient- and score-based unlearning studies~\cite{Cha:ICLR2025, Gao:ICLR2025, Ding:ICLR2025}. 
Selective Synaptic Dampening (\emph{SSD}) embodies these advantages: it compares per-parameter importance for the forget set versus overall data and applies one-shot dampening only to selected parameters, achieving strong unlearning with minimal compute and memory traffic~\cite{Foster:AAAI2023}. 
This intuition aligns with class-conditional attributions showing heterogeneous parameter contributions across classes~\cite{Selvaraju:ICCV2017}.

Despite its promise, SSD—and existing unlearning methods more broadly—remain misaligned with edge realities. First, most techniques treat layers uniformly, even though fine-grained, class-specific features concentrate in later (back-end) layers across both Convolutional Neural Networks (\emph{CNNs}) and Vision Transformers (\emph{ViTs})~\cite{Yosinski:NIPS2014, Raghu:NIPS2021}. Second, layer-agnostic hyperparameters create a tension between sufficiently suppressing back-end, class-specific detail and preserving front-end, general features, which can destabilize accuracy on the retain set. Third, the literature has not yet brought the \emph{adaptive/selective} mindset—central to edge inference~\cite{Korol:DATE2023, Motetti:DATE2024}—into the unlearning procedure itself. As a result, unnecessary computation persists, and opportunities to save energy are left on the table.

To close these gaps, we present \textbf{FiCABU}—\emph{Fisher-based Context-Adaptive Balanced Unlearning}—and a SW–HW co-designed processor for on-device unlearning (\refFigure{fig:intro2}). Our approach embraces adaptive processing to remove useless work and balances layer-wise edits to protect retained performance:

\begin{figure}
    \centering
    \vskip -2pt
    \includegraphics[width=0.84\columnwidth]{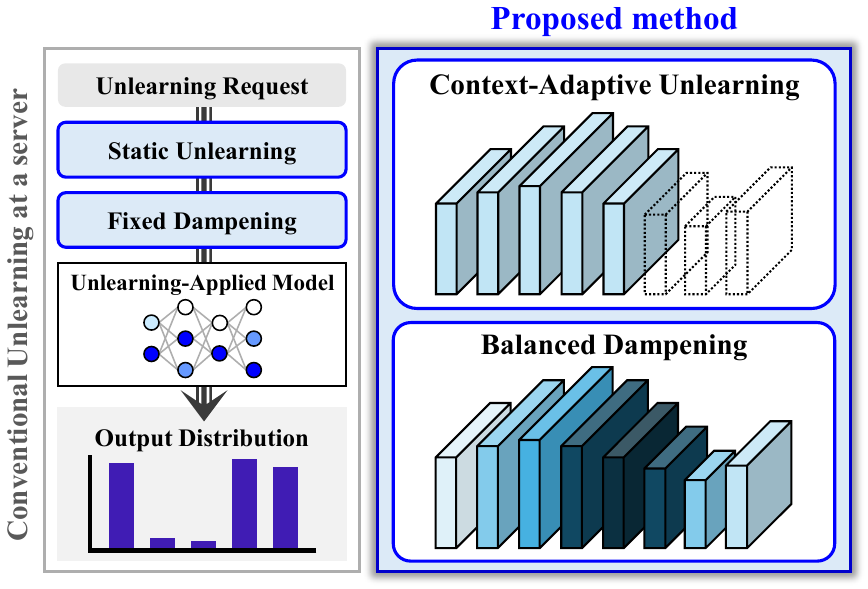}
    \vskip -4pt
\caption{\small Overview of FiCABU: context-adaptive unlearning, balanced dampening, and edge-oriented implementation.}
    \label{fig:intro2}
\end{figure} 

\begin{enumerate}[leftmargin=*,nosep]
\item \textbf{Context-Adaptive Unlearning.}
Starting from the back-end, FiCABU applies unlearning layer by layer and evaluates forget accuracy at user-defined checkpoints. Once the target unlearning level (random-guess accuracy for the task) is reached, FiCABU halts further updates on front-end layers. To minimize verification overhead, activations from the initial forward pass are cached and reused for partial inference at checkpoints. This early-stop mechanism reduces Multiply–Accumulate operations (\emph{MACs})—a hardware-relevant proxy for computation~\cite{Sandler:CVPR2018}—by an average of $87.52\%$.
\item \textbf{Balanced Dampening.}
Instead of fixed, layer-agnostic hyperparameters, FiCABU scales selection and dampening strength by a depth-dependent schedule, assigning stronger edits to back-end layers (where class-specific detail resides) while preserving front-end representations. In our experiments, this maintains unlearning efficacy while improving retain-set accuracy relative to uniform dampening.


\item \textbf{Edge Unlearning Processor.}
We realize FiCABU as a \emph{full-custom RISC-V edge AI processor} integrating a streaming General Matrix–Matrix Multiplication (\emph{GEMM}) backbone with specialized IPs for importance estimation and dampening; these stages execute concurrently within edge budgets. The processor is implemented at RTL, prototyped on a 50\,MHz Kintex-7 FPGA, and synthesized in 45\,nm for power evaluation.

\end{enumerate}


Experimental results across CIFAR-20 (ResNet-18, ViT) and PinsFaceRecognition (ResNet-18) show that FiCABU achieves random-guess forget accuracy while improving retain preservation via (i) back-end-first \emph{Context-Adaptive Unlearning} and (ii) depth-aware \emph{Balanced Dampening}. Representative figures include MACs reduced by 87.52\% on ResNet-18/CIFAR-20 and by 71.03\% on ViT/CIFAR-20 (avg.). On our INT8 hardware prototype, system energy drops to 6.48\% (CIFAR-20) and 0.13\% (PinsFaceRecognition) of the SSD baseline. Taken together, these results demonstrate that FiCABU’s method–processor co-design makes adaptive, depth-aware unlearning practical and energy-efficient on resource-constrained edge AI processors.

\section{SSD: Method and Limitations}

SSD~\cite{Foster:AAAI2023} is a recent unlearning method that removes the influence of designated data without retraining. By leveraging the diagonal approximation of the FIM and applying a one-shot dampening only to parameters strongly associated with the forget set, SSD achieves effective unlearning with substantially reduced computation and energy, making it attractive for resource-constrained edge scenarios.

As a standard formulation in machine unlearning, the full training dataset $\mathcal{D}$ is partitioned into a forget set $\mathcal{D}_f$ and a retain set $\mathcal{D}_r$, where accuracies measured on $\mathcal{D}_f$ and $\mathcal{D}_r$ are termed \emph{forget accuracy} and \emph{retain accuracy}, respectively:
\begin{align}
\mathcal{D}_f \subset \mathcal{D}, \qquad
\mathcal{D}_r = \mathcal{D} \setminus \mathcal{D}_f .
\end{align}

Parameters contribute heterogeneously to class predictions: the gradient response of a parameter to inputs from one class can differ markedly from its response to other classes~\cite{Selvaraju:ICCV2017}. 
This observation supports unlearning via \emph{selective} parameter edits guided by importance scores computed on $\mathcal{D}_f$. 
In retraining-free unlearning, the diagonal Fisher is widely used because squared first-order gradients provide a lightweight sensitivity proxy and often suffice without constructing the full matrix~\cite{Shi:ECCV2025}. 
The per-parameter diagonal Fisher on $\mathcal{D}_f$ is
\begin{align}
[\mathbb{F}_{\mathcal{D}_f}]_{ii}
=
\mathbb{E}\!\left[
\left(
\frac{\partial \ln p(\mathcal{D}_f\mid \theta)}{\partial \theta_i}
\right)^2
\right],
\quad \forall i \in [1, |\theta|],
\label{eq:diag_fisher}
\end{align}
and we denote the importance as $\mathbb{I}_{\mathcal{D}_f,i} \triangleq [\mathbb{F}_{\mathcal{D}_f}]_{ii}$. 
Because this requires only a single forward pass and the associated gradient computation to obtain the needed sensitivities, this avoids repeated full-dataset passes and hundreds of update steps, which is critical under edge compute and memory budgets.

SSD proceeds in two steps: \emph{selection} chooses parameters to modify based on their relative importance, and \emph{dampening} scales the chosen parameters by a factor $\beta \in (0,1]$. 
Simply pruning parameters by thresholding importance at zero would reduce forget accuracy but severely harm retain accuracy and discard graded control over parameter impact. 
Instead, SSD compares importance scores on the forget set against those on the overall data and applies \emph{dampening} rather than hard pruning:
\begin{align}
\theta_i =
\begin{cases}
\beta \theta_i, & \text{if } \mathbb{I}_{\mathcal{D}_f,i} > \alpha \,\mathbb{I}_{\mathcal{D},i} \\
\theta_i, & \text{if } \mathbb{I}_{\mathcal{D}_f,i} \leq \alpha \,\mathbb{I}_{\mathcal{D},i}
\end{cases}
\quad \forall i \in [1, |\theta|],
\label{eq:select}
\end{align}
where the threshold $\alpha$ controls which parameters are selected for modification. The dampening strength is
\begin{align}
\beta = \min\!\left(\frac{\lambda \,\mathbb{I}_{\mathcal{D},i}}{\mathbb{I}_{\mathcal{D}_f,i}}, \, 1\right),
\label{eq:dampening}
\end{align}
where $\lambda$ weighs the relative importance of $\mathcal{D}$ versus $\mathcal{D}_f$; smaller $\lambda$ emphasizes $\mathcal{D}_f$ and induces stronger dampening.

In the selection step of \refEqn{eq:select}, SSD uses $\mathbb{I}_{\mathcal{D}}$ rather than $\mathbb{I}_{\mathcal{D}_r}$ to avoid recomputing retain-set importance scores whenever a new forget request slightly perturbs $\mathcal{D}_r$. Since $\mathbb{I}_{\mathcal{D}}$ can be computed once after training and stored, large-scale access to the original training data is unnecessary during unlearning. Moreover, because typically $|\mathcal{D}_f|\!\ll\!|\mathcal{D}|$ and only about $1\%$ of parameters are modified per unlearning event~\cite{Foster:AAAI2023}, the difference between $\mathbb{I}_{\mathcal{D}}$ and $\mathbb{I}_{\mathcal{D}_r}$ is practically negligible.

\begin{figure}
    \centering
    \vskip -6pt
    \includegraphics[width=0.96\columnwidth]{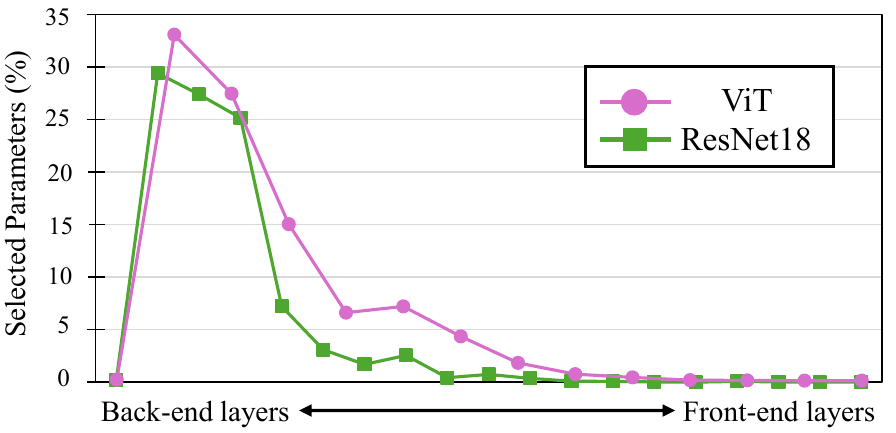}
    \vskip -4pt
	\caption{\small Layer-wise distribution of selected parameters for ResNet-18 (RN) and ViT, highlighting concentration in back-end layers.}
    \label{fig:motivation}
\end{figure}

Empirically, SSD can reduce forget accuracy to random-guess levels while maintaining performance comparable to retraining-based approaches. At that operating point, two limitations remain relevant for edge deployment. 
{First}, selection relies on a single threshold applied uniformly across all parameters, without explicitly prioritizing back-end layers where fine-grained, class-specific information is concentrated. 
{Second}, hyperparameters are layer-agnostic: despite targeting parameters that encode the detailed traits of $\mathcal{D}_f$, the dampening schedule does not adapt to depth, leaving retain accuracy sensitive to over-/under-editing across layers.


In this work, we take as our operational point the setting where SSD reduces forget accuracy to random-guess levels, which we regard as the practical target for on-device unlearning.  
We therefore consider classes that satisfy this criterion. 
Our baseline SSD configuration follows the original setup on CIFAR-20 \cite{CIFAR100}: for ResNet-18~\cite{He:CVPR2016}, $(\alpha,\lambda)=(10,1)$; for ViT~\cite{dosovitskiy:arxiv2021}, $(\alpha,\lambda)=(25,1)$. For PinsFaceRecognition\cite{pinsface}, where inter-class similarity is relatively high, we adopt $(\alpha,\lambda)=(50,0.1)$ as in~\cite{Foster:AAAI2023}. 
The batch size for the forget class is fixed at $N=64$ across all cases. 
All subsequent results build on these settings unless otherwise specified.

\section{FiCABU: Proposed Method}
The proposed {FiCABU} method combines two complementary techniques—\emph{Context-Adaptive Unlearning} and \emph{Balanced Dampening}. 
The former cuts computation by avoiding unnecessary edits on front-end layers, and the latter stabilizes retain-set performance by assigning depth-aware unlearning strength.

\subsection{Context-Adaptive Unlearning}

Empirical evidence indicates that fine-grained, class-specific features concentrate in later (back-end) layers. 
This trend is evident in \refFigure{fig:motivation}, which shows the layer-wise distribution of parameters selected during unlearning for ResNet-18 (RN) and ViT; aside from the final fully connected classifier (which has fewer parameters than convolutional layers), a clear depth-dependent pattern emerges. 
Unless otherwise noted, we index layers from the back-end (near the classifier) to the front-end (near the input): $l{=}1$ denotes the last/output layer and $l{=}L$ the first/input layer.

\DontPrintSemicolon
\begin{algorithm2e}[t]
  \caption{Context-Adaptive Unlearning}
  \label{alg:cau}
  \footnotesize
\DontPrintSemicolon
\SetKw{KwDownTo}{down to}
\SetKw{KwTo}{to}
\SetKwFunction{FCAU}{Context\_Adaptive\_Unlearning}
\SetKwFunction{FPI}{partial\_inference}
\SetKwProg{Fn}{Function}{:}{}
\SetKwInOut{KwIn}{Input}
\SetKwInOut{KwOut}{Output}

\KwIn{Pre-trained parameters $\theta$; forget mini-batch $\mathcal{D}_f$ of size $N$;
      checkpoint set $\mathcal{C}\subseteq\{1,\dots,L\}$; target $\tau$;
      stored global importance $\{\mathbb{I}_{\mathcal{D},i}\}$;
      SSD hyperparameters $(\alpha,\lambda)$.}
\KwOut{Updated parameters $\tilde\theta$.}

\tcp{Step 0: one forward pass on $\mathcal{D}_f$ and cache \emph{inputs to layer $l$} at checkpoints}
\For{$n=1$ \KwTo $N$}{
  run \texttt{forward} on sample $n$ and cache \texttt{activation[$l,n$]} for each $l\in\mathcal{C}$\;
  \tcp*{\mbox{\texttt{activation[$l,n$]} is the input tensor to $l$}}
}

\Fn{\FCAU{$\mathcal{D}_f$}}{
  \For{$l=1$ \KwTo $L$}{
     \tcp*{\mbox{{Layer update (SSD)} on layer $l$ ((\ref{eq:select}), (\ref{eq:dampening}))}}
    \tcp{Diagonal Fisher on $\mathcal{D}_f$: $\mathbb{I}_{\mathcal{D}_f,i} \leftarrow [\mathbb{F}_{\mathcal{D}_f}]_{ii}$}
    compute $\mathbb{I}_{\mathcal{D}_f,i}$ for all params in layer $l$;\;
    mark $\mathrm{Sel}_l=\{\,i:\mathbb{I}_{\mathcal{D}_f,i}>\alpha\,\mathbb{I}_{\mathcal{D},i}\,\}$ \tcp*{selection}
    \For{$i\in\mathrm{Sel}_l$}{
      $\beta_i \leftarrow \min\!\big(\lambda\,\mathbb{I}_{\mathcal{D},i}/\mathbb{I}_{\mathcal{D}_f,i},\,1\big)$;\;
      $\theta_i \leftarrow \beta_i\,\theta_i$ \tcp*{in-place dampening}
    }
    \If{$l\in\mathcal{C}$}{
      $A_{\mathrm{forget}} \gets$ \FPI{$\mathcal{D}_f$, $l$}\;
	\If{$A_{\mathrm{forget}} \le \tau$}{\textbf{break} \tcp*{leave $l{+}1,\dots,L$ untouched}}
    }
  }
  \Return $\tilde\theta \leftarrow \theta$\;
}

\Fn{\FPI{$\mathcal{D}_f$, $l$}}{
  $A_{\mathrm{sum}} \gets 0$\;
  \For{$n=1$ \KwTo $N$}{
    \texttt{act\_current} $\gets$ \texttt{activation[$l,n$]}\;
    $A_{\mathrm{temp}} \gets$ \texttt{forward}$(\texttt{act\_current},\, \texttt{from}=l,\, \texttt{to}=1)$ \tcp*{\mbox{partial inference $l\!\to\!1$ (toward back-end)}}
    $A_{\mathrm{sum}} \gets A_{\mathrm{sum}} + A_{\mathrm{temp}}$\;
  }
  \Return $\tfrac{A_{\mathrm{sum}}}{N}$ \tcp*{batch-mean forget accuracy}
}

\end{algorithm2e}

Building on this observation, we introduce \emph{Context-Adaptive Unlearning}, an edge-friendly procedure that (i) evaluates the forget objective at intermediate checkpoints and (ii) halts further unlearning once the target is achieved, thereby saving computation on remaining front-end layers. 
However, verifying intermediate performance normally requires launching new forward passes, which can incur nontrivial overhead. 
To mitigate this, we cache activations from the initial forward pass and reuse them for partial inference when evaluating checkpoints.

Concretely, SSD performs a single forward pass on $\mathcal{D}_f$, then iterates from the back-end toward the front-end, computing diagonal-Fisher importance and applying dampening layer by layer. 
This sequential process can waste work on front-end layers once sufficient forgetting has already been achieved. 
Our procedure introduces a \emph{checkpoint set} $\mathcal{C}$ along depth, as detailed in \refAlgo{alg:cau}. 
Let layers be indexed $l=1,\dots,L$ from back-end to front-end.
Starting at $l=1$ and increasing toward $L$, \texttt{Context\_Adaptive\_Unlearning} computes $\mathbb{I}_{\mathcal{D}_f,i}$ for each layer, compares it with $\mathbb{I}_{\mathcal{D},i}$, and applies in-place dampening.
If $l\in\mathcal{C}$, the routine calls \texttt{partial\_inference} using cached \texttt{activation[$l,n$]} as inputs and runs only layers $l\!\to\!1$ current layer down to the back-end) to compute the batch-mean forget accuracy $A_{\mathrm{forget}}$. 
If $A_{\mathrm{forget}}\le \tau$ (the target random-guess level), unlearning stops and layers $l{+}1,\dots,L$ are left untouched; otherwise, the loop continues with $l\!+\!1$. 
This design (i) reduces MACs by avoiding noncontributory edits, (ii) prevents unnecessary parameter changes that could harm $\mathcal{D}_r$ performance, and (iii) allows checkpoint placement to be tuned per model and budget.
For consistency, we adopt the vanilla SSD hyperparameters $(\alpha,\lambda)$ here; a depth-aware variant will be introduced in \refSection{sec:bd}.


\refTable{table:cau_main} summarizes results. 
For CIFAR-20 (\refTable{table:cau_main}-a), we compare a pre-trained {baseline model} (no unlearning applied), SSD, and our method on RN and ViT—reporting individual results for \emph{Rocket} and \emph{Mushroom (MR)} and the average across remaining classes. 
For PinsFaceRecognition (\refTable{table:cau_main}-b), we report class-averaged results on RN. 
Checkpoints are placed at the first and last layers for both models; additionally, for RN, we insert a checkpoint every four of the 16 convolutional layers, and for ViT every three of the 12 encoder layers. 
Metrics include retain-set accuracy on $\mathcal{D}_r$, forget-set accuracy on $\mathcal{D}_f$ (desired to reach random-guess level), Membership Inference Attack (MIA) accuracy (lower is better), and the number of MACs. 
MACs are reported relative to SSD (normalized to $100\%$), and include the overhead of checkpoint evaluation.

\begin{table}[t]
\vskip -2pt
    \centering
\caption{\small Context-Adaptive Unlearning vs.\ baseline (pre-trained model without unlearning) and SSD. Rows for $\mathcal{D}_r$ and $\mathcal{D}_f$ report accuracy on the respective sets. All values are in percent [\%].}   
    \label{table:cau_main}

    \begin{subtable}{\columnwidth}
        \centering
\caption{\small CIFAR-20 results: unlearning performance of RN and ViT on {Rocket} and {MR} classes, and the average across the remaining classes.}
       \vskip -2pt
        \resizebox{\linewidth}{!}{\renewcommand{\arraystretch}{1.12}%

\resizebox{\columnwidth}{!}{%
\begin{tabular}{c|c|ccc|ccc}
\hline
 &  & \multicolumn{3}{c|}{\textbf{RN}} & \multicolumn{3}{c}{\textbf{ViT}} \\
\cline{3-8}
\textbf{Class} & \textbf{Metric} & Baseline & SSD & \textbf{Ours} & Baseline & SSD & \textbf{Ours} \\
\hline
\multirow{4}{*}{Rocket}
 & $\mathcal{D}_r$            & 96.95 & 96.14 & \textbf{96.50} & 94.94 & 94.91 & \textbf{94.93} \\
 & $\mathcal{D}_f$            & 97.60 &  3.80 & \textbf{3.80}  & 95.00 &  3.20 & \textbf{5.00} \\
 & MIA                        & 82.00 &  9.20 & \textbf{5.40}  & 65.40 & 20.40 & \textbf{20.20} \\
 & MACs                       & --    & 100   & \textbf{32.59} & --    & 100   & \textbf{74.47} \\
\hline\hline
\multirow{4}{*}{MR}
 & $\mathcal{D}_r$            & 96.94 & 95.35 & \textbf{95.46} & 94.50 & 94.32 & \textbf{94.33} \\
 & $\mathcal{D}_f$            & 98.40 &  1.20 & \textbf{3.20}  & 96.80 &  1.80 & \textbf{3.20} \\
 & MIA                        & 86.20 &  6.00 & \textbf{6.00}  & 61.80 &  9.60 & \textbf{5.60} \\
 & MACs                       & --    & 100   & \textbf{32.59} & --    & 100   & \textbf{33.10} \\
\hline\hline
\multirow{4}{*}{Avg.}
 & $\mathcal{D}_r$            & 96.95 & 92.28 & \textbf{93.07} & 93.72 & 91.22 & \textbf{91.86} \\
 & $\mathcal{D}_f$            & 97.31 &  0.71 & \textbf{0.86}  & 83.25 &  1.26 & \textbf{2.98} \\
 & MIA                        & 79.56 &  9.44 & \textbf{5.44}  & 49.33 & 25.60 & \textbf{17.42} \\
 & MACs                       & --    & 100   & \textbf{12.48} & --    & 100   & \textbf{28.97} \\
\hline
\end{tabular}
}

}
        \vspace{2pt}
        \label{table:cau_CIFAR}
    \end{subtable}

    \vspace{5pt}

    \begin{subtable}{\columnwidth}
        \centering
\caption{\small PinsFaceRecognition results: unlearning performance of RN, reported as the average across classes.}
                 \vskip -2pt
        \resizebox{0.9\columnwidth}{!}{{%
\renewcommand{\arraystretch}{1.12}%
\setlength{\arrayrulewidth}{0.35pt} 
\hspace{1em}
\centering
\label{table:rn_results}
\vspace{-4pt}
\begin{tabular}{@{\hspace{0.6em}} c|l|c|c|>{\bfseries}c @{\hspace{0.6em}}}
\hline
~~~~~~~Model~~~~~~~ & Metric & Baseline & SSD & Ours \\
\hline
\multirow{4}{*}{\textbf{RN}}
  & $\mathcal{D}_r$ & 98.11 & 97.92 & 98.08 \\
  & $\mathcal{D}_f$ & 97.63 & 0.00  & 0.00 \\
  & MIA             & 41.61 & 0.46  & 0.11 \\
  & MACs        &   --   & 100 & 0.00137 \\
\hline
\end{tabular}
\hspace{1em}
\vspace{1em}
}
}
        \vspace{2pt}
        \label{table:cau_pins}
    \end{subtable}
\end{table}

Overall, Context-Adaptive Unlearning matches SSD on retain accuracy and MIA while reaching random-guess forget accuracy (5\% on CIFAR-20; 1\% on PinsFaceRecognition) across all evaluated cases. 
Using MACs that include checkpoint overhead, RN on CIFAR-20 shows identical early-stop checkpoints for \emph{Rocket} and \emph{Mushroom} with a $67.41\%$ reduction, and an average reduction of $87.52\%$ over other classes.
ViT achieves $25.53\%$ ({Rocket}), $66.90\%$ ({MR}), and $71.03\%$ on average for the rest. 
For PinsFaceRecognition with RN, the average MACs reduction reaches about $99.9\%$, which can plausibly be attributed to the higher inter-class similarity in face data, leading to a stronger concentration of discriminative detail in later layers.
In summary, the method substantially reduces computation while preserving unlearning quality—an essential property for an edge-targeted unlearning processor.

  \vspace{-4pt}
\subsection{Balanced Dampening}\label{sec:bd}

While Context-Adaptive Unlearning removes unnecessary edits by early stopping, it does not address the second limitation of SSD: layer-agnostic hyperparameters. 
In \refEqn{eq:select}, the threshold $\alpha$ determines selection (relative emphasis of $\mathcal{D}_f$ vs.\ $\mathcal{D}$), and in \refEqn{eq:dampening}, $\lambda$ controls dampening strength. 
Smaller values induce stronger forgetting. 

\begin{figure}
    \centering
    \vskip -6pt
    \includegraphics[width=0.98\columnwidth]{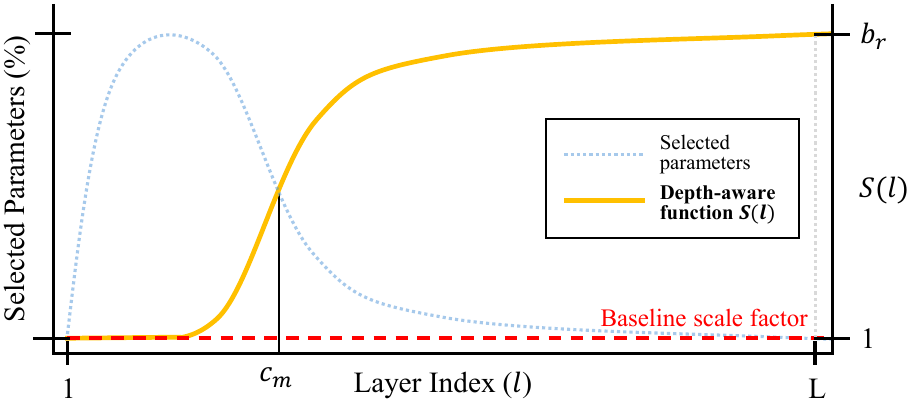}
    \vskip -4pt
    \caption{\small Baseline uniform scaling vs.\ proposed sigmoid-based profile; $S(l)$ is {smaller} at the back-end and {larger} at the front-end.}
    \label{fig:profile}
     \vskip 5pt
\end{figure}

\begin{table}[t]
    \centering
    \caption{\small Balanced Dampening vs. baseline and SSD. All values are in percent [\%].} 
    \label{table:bd_main}

    \begin{subtable}{\columnwidth}
            \caption{\small CIFAR-20: Results for RN and ViT on Rocket, MR, and the average across remaining classes.}
         \vskip -4pt
        \centering
        \resizebox{1\columnwidth}{!}{\renewcommand{\arraystretch}{1.12}%

\resizebox{\columnwidth}{!}{%
\begin{tabular}{c|c|ccc|ccc}
\hline
 &  & \multicolumn{3}{c|}{\textbf{RN}} & \multicolumn{3}{c}{\textbf{ViT}} \\
\cline{3-8}
\textbf{Class} & \textbf{Metric} & Baseline & SSD & \textbf{Ours} & Baseline & SSD & \textbf{Ours} \\
\hline
\multirow{4}{*}{Rocket}
 & $\mathcal{D}_r$            & 96.95 & 96.14 & \textbf{96.25} & 94.94 & 94.91 & \textbf{94.92} \\
 & $\mathcal{D}_f$            & 97.60 &  3.80 & \textbf{3.80}  & 95.00 &  3.20 & \textbf{3.60} \\
 & $\Delta\mathcal{D}_r$      &   --  &  0.81 & \textbf{0.70}  & --    &  0.03 & \textbf{0.02} \\
 & RPR                        &   --  &   --  & \textbf{13.58} &   --  &  --   & \textbf{33.33} \\
\hline\hline 
\multirow{4}{*}{MR}
 & $\mathcal{D}_r$            & 96.94 & 95.35 & \textbf{95.37} & 94.50 & 94.32 & \textbf{94.35} \\
 & $\mathcal{D}_f$            & 98.40 &  1.20 & \textbf{1.20}  & 96.80 &  1.80 & \textbf{3.00} \\
 & $\Delta\mathcal{D}_r$      &   --  &  1.59 & \textbf{1.57}  &   --  &  0.18 & \textbf{0.15} \\
 & RPR                        &   --  &   --  & \textbf{1.26}  &   --  &   --  & \textbf{16.67} \\
\hline\hline 
\multirow{4}{*}{Avg.}
 & $\mathcal{D}_r$            & 96.95 & 92.28 & \textbf{92.58} & 93.72 & 91.22 & \textbf{91.61} \\
 & $\mathcal{D}_f$            & 97.31 &  0.71 & \textbf{0.73}  & 83.25 &  1.26 & \textbf{2.67} \\
 & $\Delta\mathcal{D}_r$      &   --  &  4.67 & \textbf{4.37}  &   --  & 2.50 & \textbf{2.11} \\
 & RPR                        &   --  &   --  & \textbf{6.42}  &   --  &   --  & \textbf{15.60} \\
\hline
\end{tabular}
}

}
        \vspace{2pt}
        \label{table:bd_CIFAR}
    \end{subtable}

    \vspace{6pt}

    \begin{subtable}{\columnwidth}
        \caption{\small PinsFaceRecognition: Results for RN, reported as the class average.}
         \vskip -4pt
        \centering
        \resizebox{0.8\columnwidth}{!}{\renewcommand{\arraystretch}{1.12}%
\setlength{\arrayrulewidth}{0.35pt} 

\resizebox{0.55\columnwidth}{!}{%
\begin{tabular}{c|l|c|c|c}
\hline
~~~~~~~Model~~~~~~~ & Metric & Baseline & SSD & \textbf{Ours} \\
\hline
\multirow{4}{*}{\textbf{RN}}
  & $\mathcal{D}_r$        & 98.11 & 97.92 & \textbf{97.95} \\
  & $\mathcal{D}_f$        & 97.63 & 0.00  & \textbf{0.00} \\
  & $\Delta\mathcal{D}_r$  &   --  & 0.19  & \textbf{0.16} \\
  & RPR                    &   --  &  --   & \textbf{15.79} \\
\hline
\end{tabular}
}}
        \label{table:bd_pins}
    \end{subtable}
\end{table}

We generalize these scalars into a depth-aware profile so that the {back-end} layers receive relatively stronger edits (and the {front-end} layers, weaker ones):
\begin{align}
{(\alpha,\lambda)\;\longrightarrow\; S(l)\cdot(\alpha,\lambda).}
\label{eq:balanced}
\end{align}

Motivated by reports that back-end layers encode more class-specific detail, we instantiate $S(l)$ as a sigmoid that is {smaller} at back-end layers ($l$ small) and {larger} at front-end layers ($l$ large).
As illustrated in \refFigure{fig:profile}, the baseline scale factor (red dashed) is uniform across layers, whereas the proposed $S(l)$ (yellow) {monotonically 
{increases} with $l$} (i.e., it is {smaller} near the back-end and {larger} near the front-end), {which qualitatively mirrors the observed distribution of selected parameters in reverse (blue dotted).}
The midpoint $c_m$ determines the transition region, and the bound $b_r$ (``bound for retain'') controls the {front-end} scaling level. Formally, $S(l)$ is defined as
\begin{align}
S(l) &= {1} + {({b_r}-1)}\cdot 
       \frac{\sigma(l)-\sigma({1})}{\sigma({L})-\sigma({1})},
\qquad 1 \le l \le L,
\label{eq:sigmoid}
\end{align}
where $\displaystyle \sigma(l) = \frac{1}{1+\exp(-(l-c_m))}$.
This replaces fixed scalars with a depth-aware function, enabling stronger dampening where class-specific features reside while protecting generic features in the front-end.

\refTable{table:bd_main} quantitatively compares Balanced Dampening to the baseline and SSD under the same settings as \refTable{table:cau_main}. 
For all experiments, we (i) compute the distribution of modified parameters under baseline SSD, (ii) smooth and center the sigmoid midpoint $c_m$ at the mid-value between the smoothed extrema, and (iii) set $b_r=10$. 
Metrics include forget/retain accuracy, the change in retain accuracy relative to baseline ($\Delta\mathcal{D}_r$), and the \emph{Retain Preservation Rate} (RPR), which measures how well the retain accuracy is preserved:
\begin{equation}
{\small
\begin{aligned}
\mathrm{RPR}
&= \left(1 - \frac{\Delta \mathcal{D}_r^{\mathrm{Ours}}}{\Delta \mathcal{D}_r^{\mathrm{SSD}}}\right)\times 100
\end{aligned}
}
\label{eq:rpr}
\end{equation}
where $\Delta \mathcal{D}_r^{\mathrm{SSD}}$ and $\Delta \mathcal{D}_r^{\mathrm{Ours}}$ denote the retain accuracy drops under SSD and under our method, respectively.

As reported in \refTable{table:bd_main}-a, Balanced Dampening achieves random-guess forget accuracy comparable to SSD while consistently improving retain preservation (positive RPR across all cases). On CIFAR-20, RN yields RPR of $13.58$ ({Rocket}), $1.26$ ({MR}), and $6.42$ on average across remaining classes; ViT yields $33.33$, $16.67$, and $15.60$, respectively. 
For PinsFaceRecognition (\refTable{table:bd_main}-b), RN achieves an average RPR of $15.79$. In short, depth-aware scaling maintains unlearning efficacy while better safeguarding retain-set performance. 
Because our experiments are conducted on well-converged pre-trained models, the observed RPR improvement should be regarded as a conservative lower bound. In contrast, edge deployments typically use centrally trained models but perform limited-data, short-horizon on-device updates (rather than full training), where depth sensitivity and overfitting—especially in later layers—tend to be more pronounced; under these conditions, the benefit of the proposed schedule is expected to amplify.

\begin{figure}
\vskip -8pt
\centering
\includegraphics[width=0.84\columnwidth]{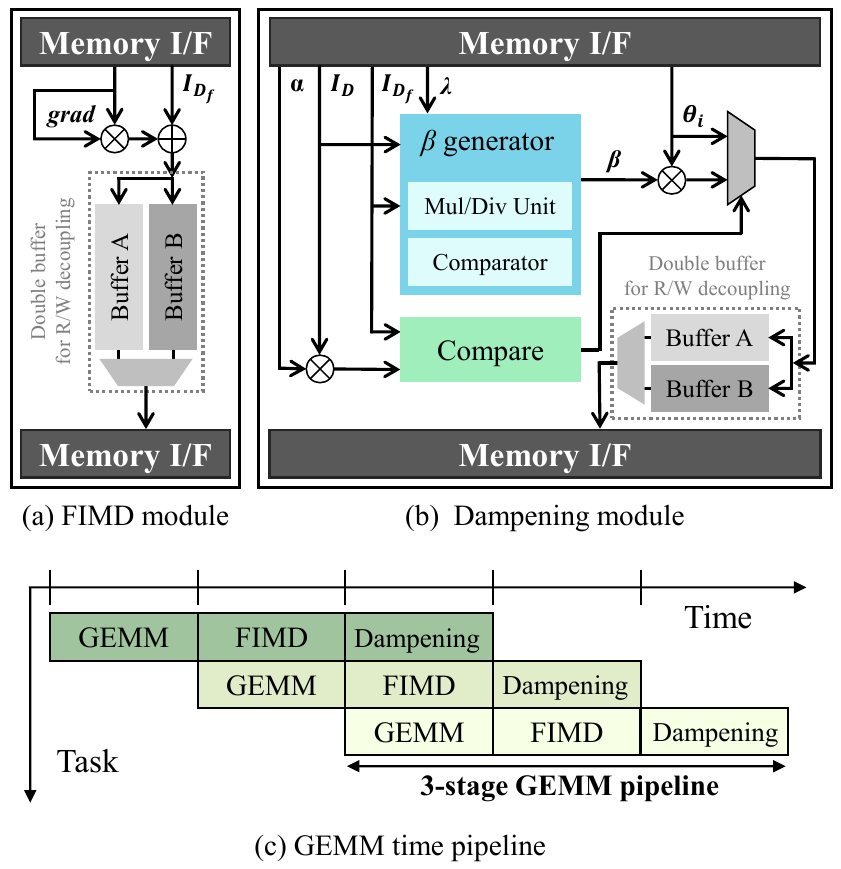}
\vskip -1pt
\caption{\small Key hardware components of FiCABU:
(a) FIMD module for diagonal Fisher estimation used by Context-Adaptive Unlearning,
(b) Dampening module implementing Balanced Dampening updates, and
(c) patch-level GEMM–FIMD–Dampening pipeline.}
\label{fig:arch}
\vskip 5pt
\end{figure}

\vspace{+4pt}
\section{Implementation and Evaluation} 

\subsection{FiCABU Processor: Design and Prototyping}
\label{sec:hardware}

To realize on-device unlearning under edge constraints, we co-design FiCABU in software and hardware and integrate them into a lightweight processor. 
Because the dominant arithmetic in our workflow is matrix multiplication (forward-pass activations, gradient calculation), we adopt a dedicated GEMM accelerator as the backbone and attach two specialized IPs for (i) diagonal-Fisher approximation and (ii) dampening, as shown in \refFigure{fig:arch}.


\textbf{GEMM-centric pipeline.}
During unlearning, the system performs a single forward pass over the forget batch to compute and cache layer-wise activations. 
Subsequent importance estimation and dampening reuse these cached activations together with back-propagated gradients, making GEMM the performance and energy hotspot. 
The GEMM engine executes matrix multiplications in fixed-size \emph{patch} (tile) units and streams operands from memory; this streaming organization allows us to align the auxiliary IPs with the patch cadence.


\textbf{Fisher Information Matrix Diagonal (FIMD) IP.}
The FIMD module (\refFigure{fig:arch}a) consumes the gradient outputs produced by the GEMM engine and implements the diagonal Fisher in \refEqn{eq:diag_fisher} by squaring elements and accumulating them across the batch dimension to obtain $\mathbb{I}_{\mathcal{D}_f}$. To avoid read/write contention when accessing large parameter/gradient buffers, FIMD uses double buffering. 
Internally, it forms a four-stage pipeline---\textsc{Load} $\rightarrow$ \textsc{Square} $\rightarrow$ \textsc{Accumulate} $\rightarrow$ \textsc{Store}---so that data transfers and arithmetic overlap. 
As a result, diagonal-Fisher computation that previously created a bottleneck when executed on the core is accelerated by $11.7\times$ on the FIMD module, hiding its latency within the GEMM patch window.


\textbf{Dampening IP.}
The Dampening module (\refFigure{fig:arch}b) realizes the selection rule in \refEqn{eq:select} and the strength schedule in \refEqn{eq:dampening}. 
For each parameter, it compares $\mathbb{I}_{\mathcal{D}_f,i}$ with $\alpha\,\mathbb{I}_{\mathcal{D},i}$, generates the per-parameter coefficient $\beta$ via a dedicated \textsc{$\beta$ Generator} when selected, and updates the value by multiplication. 
Similar to FIMD, it employs a double-buffered datapath and a five-stage pipeline---\textsc{Load} $\rightarrow$ \textsc{Compare} $\rightarrow$ \textsc{$\beta$ Calc} $\rightarrow$ \textsc{Multiply} $\rightarrow$ \textsc{Store}---interleaving memory traffic with computation. 
This organization eliminates the bottleneck of dampening observed on the core, achieving a $7.9\times$ speedup and ensuring completion within the GEMM patch window.

\begin{figure}
\vskip -8pt
\centering
\includegraphics[width=1\columnwidth]{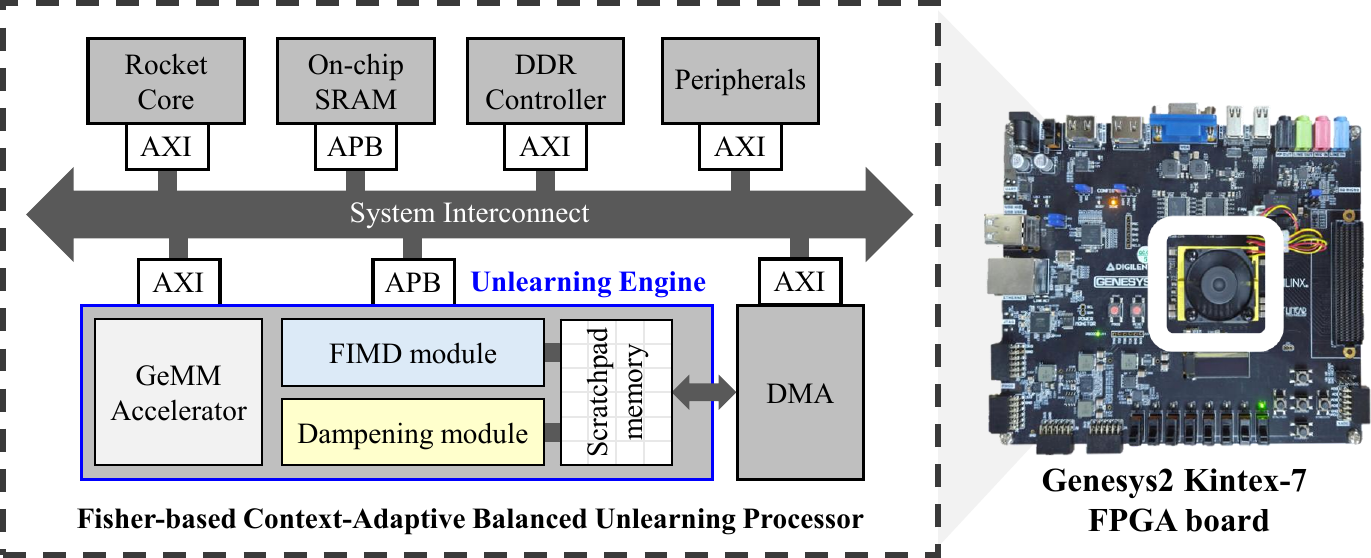}
\vskip -2pt
\caption{\small FiCABU processor architecture (left) and FPGA prototype on a Genesys2 Kintex-7 board (right).}
\label{fig:arc}
\vskip 3pt
\end{figure}

\begin{table}
\caption{\small FPGA resource utilization of the FiCABU processor and power consumption estimated from 45\,nm process technology.}
	\centering
\begin{center}
	\vskip -10pt
	 \resizebox{0.84\columnwidth}{!}{\resizebox{0.95\columnwidth}{!}{
    
\renewcommand{\arraystretch}{1.1}
\begin{tabular}{cll|c|c|c}
\Xhline{1pt}
\multicolumn{3}{c|}{{IPs}}            & {LUTs}  & {FFs} & $P_{total}$\,$(mW)$ \\ \Xhline{1.5pt}

\multicolumn{3}{l|}{\cellcolor[HTML]{EFEFEF}FiCABU Processor}  
& \cellcolor[HTML]{EFEFEF}71,535  
& \cellcolor[HTML]{EFEFEF}35,059  
& \cellcolor[HTML]{EFEFEF}185.89 \\ \hline

\multicolumn{3}{l|}{$\llcorner$ RISC-V Rocket Core} & 15,246    & 9,756 & 11.2  \\ \hline
\multicolumn{3}{l|}{$\llcorner$ SRAM}               & 354       & 653   & 1.71  \\ \hline
\multicolumn{3}{l|}{$\llcorner$ DMA}                & 1,556     & 951   & 4.07  \\ \hline
\multicolumn{3}{l|}{$\llcorner$ Peripherals}        & 4,329      & 7,562  & 5.68  \\ \hline
\multicolumn{3}{l|}{$\llcorner$ DDR Controller}     & 8,102     & 7,514 & 88.62 \\ \hline
\multicolumn{3}{l|}{$\llcorner$ System Interconnect}& 5,234     & 652 & 33.9 \\ \hline

\multicolumn{3}{l|}{$\llcorner$ \cellcolor[HTML]{F5F5F5}\textbf{Unlearning Engine}}  
& \cellcolor[HTML]{F5F5F5}\textbf{36,714}
& \cellcolor[HTML]{F5F5F5}\textbf{7,971}
& \cellcolor[HTML]{F5F5F5}\textbf{40.71} \\ \hline

& \multicolumn{2}{l|}{$\llcorner$ VTA} & 34,529 & 7,186 & 39.9 \\ \hline

\multicolumn{1}{>{\columncolor[HTML]{F5F5F5}}c}{}%
& \multicolumn{2}{>{\columncolor[HTML]{F5F5F5}}l|}{$\llcorner$ \textbf{Specialized IPs}}
& \cellcolor[HTML]{F5F5F5}\textbf{2,185}
& \cellcolor[HTML]{F5F5F5}\textbf{785}
& \cellcolor[HTML]{F5F5F5}\textbf{0.81} \\ \hline

\end{tabular}
}}
\end{center}
	\label{table:resource_power}
\vskip -2pt
\end{table}


\vspace{1pt}
\textbf{End-to-end streaming.}
By aligning the three IPs, FiCABU forms a three-stage, patch-level streaming pipeline---\textsc{GEMM} $\rightarrow$ \textsc{FIMD} $\rightarrow$ \textsc{Dampening}---that operates continuously at the GEMM patch rate (\refFigure{fig:arch}c). This organization hides the internal latency of the specialized IPs behind GEMM processing and ensures high throughput without enlarging on-chip buffers.

\vspace{1pt}
\textbf{Platform integration and prototyping.}
As shown in Fig.~\ref{fig:arc}, we implemented the full-custom FiCABU processor at the RTL level on a RISC-V platform using RISC-V eXpress (RVX), an EDA environment that leverages open-source RISC-V cores for rapid processor design and customization~\cite{Han:IoT2021,CHOI:JESTCH2024,Park:TCASI24,Lee:IoTJ25,CHOI:AEJ25}.
The system comprises a 50\,MHz RISC-V Rocket core~\cite{Rocket}, 64\,KB on-chip SRAM, a DDR controller, peripherals, AXI DMA, and a $\mu$NoC with AXI/APB interconnects. 
For GEMM, we integrate the open-source Versatile Tensor Accelerator (VTA)~\cite{vta} for rapid system bring-up and reproducibility; exploring custom minimal GEMM designs is orthogonal to our contributions.
The two specialized IPs (FIMD and Dampening) control a custom DMA that orchestrates bulk parameter transfers between main memory and the scratchpad, forming the \emph{Unlearning Engine} with the GEMM accelerator.
Unless noted otherwise, we target INT8 quantized models, consistent with common edge deployments. 
We prototype the processor on a Genesys2 Kintex-7 FPGA board and synthesize in a 45\,nm technology using FreePDK45 for power estimation.

\vspace{1pt}
\textbf{Resource and power breakdown.}
\refTable{table:resource_power} summarizes the FPGA and ASIC-level costs. 
The FPGA prototype uses 71{,}535 LUTs and 35{,}059 FFs; the Unlearning Engine accounts for 36{,}714 LUTs and 7{,}971 FFs (51.3\% and 22.7\% of the totals, respectively). In ASIC estimates using Design Compiler, the Unlearning Engine draws 40.71\,mW (21.9\% of system power), while the DDR controller and VTA consume 88.62\,mW (47.7\%) and 39.90\,mW (21.5\%), respectively. Notably, the \emph{Specialized IPs} (FIMD + Dampening) are lightweight, using only 2{,}185 LUTs (3.1\%) and 785 FFs (2.2\%), with 0.81\,mW (0.44\%) of power. Although off-chip memory and GEMM dominate the power budget---a typical characteristic of edge inference systems that store large models in DRAM---the proposed IPs add minimal overhead while enabling the streaming pipeline that sustains throughput at the GEMM rate. Putting these pieces together, the FiCABU processor sustains GEMM‑rate throughput while incurring only negligible power/area overhead for the two specialized IPs. 
In the following subsection, we evaluate the developed FiCABU processor.

\begin{table}
\vskip +5pt
\caption{\small Unlearning performance and energy savings~(\textit{ES}) of the FiCABU processor running an INT8-quantized ResNet-18 on CIFAR-20 and PinsFaceRecognition. ES is measured relative to SSD on the baseline processor. All values are in percent [\%].}
	\centering
\begin{center}
	\vskip -4pt        
	\resizebox{0.95\columnwidth}{!}{\renewcommand{\arraystretch}{1.12}
\setlength{\tabcolsep}{4pt}

\resizebox{\linewidth}{!}{%
\begin{tabular}{c|ccc|ccc}
\hline
 & \multicolumn{3}{c|}{CIFAR-20} & \multicolumn{3}{c}{ PinsFaceRecognition} \\
\cline{2-7}
\textbf{Metric} & Baseline & SSD & \textbf{FiCABU} & Baseline & SSD & \textbf{FiCABU} \\
\hline
$\mathcal{D}_r$ & 96.96 & 88.87 & \textbf{93.08} & 98.11 & 97.95 & \textbf{98.02} \\
$\mathcal{D}_f$ & 96.40 & 0.41  & \textbf{1.21}  & 97.70 & 0.00  & \textbf{0.00} \\
MACs            & --    & 100   & \textbf{11.88} & --    & 100   & \textbf{0.0014} \\
RPR             & --    & --    & \textbf{52.04} & --    & --    & \textbf{43.75} \\
$ES$             & --    & --    & \textbf{93.52} & --    & --    & \textbf{99.87} \\
\hline
\end{tabular}
}

}
\end{center}
    \label{table:4-2}
\vskip -4pt

\end{table}

\subsection{End-to-End Evaluation on the FiCABU Processor}
To evaluate FiCABU end-to-end, we implemented the FiCABU processor and also prototyped a baseline processor that comprises the same components (e.g., GEMM accelerator, Rocket core, memory subsystem) but excludes the specialized unlearning IPs. The baseline method (pre-trained model without unlearning) and SSD were executed on this baseline processor for comparison. Unlike the FP32 simulations in Section~III, all experiments here used INT8 ResNet-18 models to reflect hardware deployment; the remaining settings followed Section~III.

\refTable{table:4-2} reports measured performance. Across both datasets, the FiCABU processor attains random-guess forget accuracy while reducing MACs by 88.11\% (CIFAR-20) and 99.998\% (PinsFaceRecognition) relative to SSD on the baseline processor. This confirms on hardware what Section~III suggested in simulation: \emph{Context-Adaptive Unlearning} can stop at back-end layers and still achieve the target forgetting, thereby realizing the context-adaptive processing capability essential for the edge. Retain accuracy also improves, yielding positive RPR in both cases; this indicates that applying depth-aware dampening per layer effectively preserves general performance on hardware, as in simulation. Compared to \refTable{table:bd_main}, the RPR gains are stronger here because the evaluation combines \emph{Balanced Dampening} with \emph{Context-Adaptive Unlearning}, and the latter halts edits on front-end layers, preventing excessive dampening of general features.

Energy results are also summarized in \refTable{table:4-2}. In addition to substantial MAC reductions (and the resulting runtime decrease), hardware assistance from the specialized IPs and patch-level pipelining, which consume only 0.44\% of system power as reported in \refTable{table:resource_power}, reduces system energy to 6.48\% (CIFAR-20) and 0.13\% (PinsFaceRecognition) of the SSD baseline. In summary, FiCABU preserves unlearning quality while driving energy to extremely low levels, firmly demonstrating that machine unlearning can be both feasible and efficient on resource-constrained edge AI processors.

\section{Conclusion}
We introduced \emph{FiCABU}, a SW–HW co-design that enables practical unlearning on edge AI processors. 
FiCABU combines \emph{Context-Adaptive Unlearning}, which begins edits from back-end layers and halts once the target forgetting is reached, with \emph{Balanced Dampening}, which scales dampening strength by depth. 
These methods are realized in a full RTL design of an edge AI processor that integrates two lightweight IPs for Fisher estimation and dampening into a GEMM-centric streaming pipeline, validated on a FPGA prototype and synthesized in 45\,nm for power analysis.
Experiments on CIFAR-20 and PinsFaceRecognition show that FiCABU achieves random-guess forget accuracy while matching SSD on retain accuracy, with computation reduced by up to $87.52\%$ (ResNet-18) and $71.03\%$ (ViT). 
On the INT8 hardware prototype, FiCABU preserves these benefits and further improves retain preservation, while system-level energy drops to $6.48\%$ (CIFAR-20) and $0.13\%$ (PinsFaceRecognition) of the SSD baseline. 
In sum, FiCABU demonstrates that back-end–first, depth-aware unlearning can be made practical and efficient for resource-constrained edge AI devices.


%
%

\bibliographystyle{IEEEtran}
\bibliography{reference}

\end{document}